\documentclass[11pt, a4paper] {scrartcl}

\usepackage{amsmath}
\usepackage[margin=1.0in]{geometry}

\usepackage{times}
\usepackage{url}
\usepackage{latexsym}

\title{A Generalized Language Model as the Combination of Skipped $n$-grams and Modified Kneser-Ney Smoothing}

\author{
  Rene Pickhardt, Thomas Gottron, Martin K\"orner, Steffen Staab \\
\small{Institute for Web Science and Technologies}\\
\small{University of Koblenz-Landau, Germany}\\
\small{{\tt \{rpickhardt,gottron,mkoerner,staab\}@uni-koblenz.de}} \\\and
  Paul Georg Wagner and Till Speicher \\
\small{Typology GbR} \\
\small{{\tt mail@typology.de}}
}
\date{\today}

\usepackage{url}
\usepackage{amssymb}
\usepackage{epsfig}
\usepackage{subfigure}
\usepackage{amsmath}
\usepackage{flushend}
\usepackage{tabularx}
\newcolumntype{Y}{>{\raggedright\arraybackslash}X}
\usepackage{booktabs}
\usepackage{amssymb}

\usepackage{tikz}

\usepackage{xspace}

\newcommand\copyrighttext{%
  \footnotesize The Copyright of this work is owned by the Association for Computational Linguistics (ACL). However, each of the authors and the employers for whom the work was performed reserve all other rights, specifically including the following: ... (4) The right to make copies of the work for internal distribution within the author's organization and for external distribution as a preprint, reprint, technical report, or related class of document. The original paper is to appear in: ACL 2014: Proceedings of the 52nd Annual Meeting of the Association for Computational Linguistics}
\newcommand\copyrightnotice{%
\begin{tikzpicture}[remember picture,overlay]
\node[anchor=south,yshift=10pt] at (current page.south) {\fbox{\parbox{\dimexpr\textwidth-\fboxsep-\fboxrule\relax}{\copyrighttext}}};
\end{tikzpicture}%
}

\newcommand{\skp}[1]{\partial_{#1}}

\begin{document}
\maketitle 

\copyrightnotice

\begin{abstract}
We introduce a novel approach for building language models based on a systematic, recursive exploration of skip $n$-gram models which are interpolated using modified Kneser-Ney smoothing.
Our approach generalizes language models as it contains the classical interpolation with lower order models as a special case.
In this paper we motivate, formalize and present our approach.
In an extensive empirical experiment over English text corpora we demonstrate that our generalized language models lead to a substantial reduction of perplexity between $3.1\%$ and $12.7\%$ in comparison to traditional language models using modified Kneser-Ney smoothing.
Furthermore, we investigate the behaviour over three other languages and a domain specific corpus where we observed consistent improvements.
Finally, we also show that the strength of our approach lies in its ability to cope in particular with sparse training data.
Using a very small training data set of only $736$ KB text we yield improvements of even $25.7\%$ reduction of perplexity.
\end{abstract}
\enlargethispage{-5\baselineskip}
\section{Introduction motivation}
Language Models are a probabilistic approach for predicting the occurrence of a sequence of words.
They are used in many applications, e.g.\ word prediction~\cite{P:HLT:2005:BickelHS}, speech recognition~\cite{rabiner1993fundamentals}, machine translation~\cite{brown1990statistical}, or spelling correction~\cite{mays1991context}.
The task language models attempt to solve is the estimation of a probability of a given sequence of words $w_{1}^{l} = w_1,\dots,w_l$.
The probability $P(w_{1}^{l})$ of this sequence can be broken down into a product of conditional probabilities:

{\small
\begin{align}
    P(w_{1}^{l})=&P(w_1) \cdot P(w_2|w_1) \cdot \ldots \cdot P(w_l| w_1\cdots w_{l-1}) 
    =&\prod_{i=1}^{l}P(w_i|w_1\cdots w_{i-1}) \label{eq:probseq}
\end{align}
}%

Because of combinatorial explosion and data sparsity, it is very difficult to reliably estimate the probabilities that are conditioned on a longer subsequence. 
Therefore, by making a Markov assumption the true probability of a word sequence is only approximated by restricting conditional probabilities to depend only on a  local context $w_{i-n+1}^{i-1}$ of  $n-1$ preceding words rather than the full sequence $w^{i-1}_1$.
The challenge in the construction of language models is to provide reliable estimators for the conditional probabilities.
While the estimators can be learnt---using, e.g., a maximum likelihood estimator over $n$-grams obtained from training data---the obtained values are not very reliable for events which may have been observed only a few times or not at all in the training data.

Smoothing is a standard technique to overcome this data sparsity problem. 
Various smoothing approaches have been developed and applied in the context of language models.
Chen and Goodman~\cite{J:CSL:1999:ChenG} introduced modified Kneser-Ney Smoothing, which up to now has been considered the state-of-the-art method for language modelling over the last 15 years.
Modified Kneser-Ney Smoothing is an interpolating method which combines the estimated conditional probabilities $P(w_{i}|w_{i-n+1}^{i-1})$ recursively with lower order models involving a shorter local context  $w_{i-n+2}^{i-1}$ and their estimate for $P(w_{i}|w_{i-n+2}^{i-1})$.
The motivation for using lower order models is that shorter contexts may be observed more often and, thus, suffer less from data sparsity.
However, a single rare word towards the end of the local context will always cause the context to be observed rarely in the training data and hence will lead to an unreliable estimation.

Because of Zipfian word distributions, most words occur very rarely and hence their true probability of occurrence may be estimated only very poorly. 
One word that appears at the end of a local context $w^{i-1}_{i-n+1}$ and for which only a poor approximation exists may adversely affect the conditional probabilities in language models of all lengths --- leading to severe errors even for smoothed language models.
Thus, the idea motivating our approach is to involve several lower order models which systematically leave out one position in the context (one may think of replacing the affected word in the context with a wildcard) instead of shortening the sequence only by one word at the beginning. 

This concept of introducing gaps in $n$-grams is referred to as skip $n$-grams~\cite{J:CLS:1994:NeyEK,J:CSL:1993:HuangFHM}. 
Among other techniques, skip $n$-grams have also been considered as an approach to overcome problems of data sparsity~\cite{Tech:2001:Goodman}.
However, to best of our knowledge, language models making use of skip $n$-grams models have never been investigated to their full extent and over different levels of lower order models.
Our approach differs as we consider all possible combinations of gaps in a local context and interpolate the higher order model with all possible lower order models derived from adding gaps in all different ways.

In this paper we make the following contributions:
\begin{enumerate}
\item We provide a framework for using modified Kneser-Ney smoothing in combination with a systematic exploration of lower order models based on skip $n$-grams.
\item We show how our novel approach can indeed easily be interpreted as a generalized version of the current state-of-the-art language models.
\item We present a large scale empirical analysis of our generalized language models on eight data sets spanning four different languages, namely, a wikipedia-based text corpus and the JRC-Acquis corpus of legislative texts.
\item We empirically observe that introducing skip $n$-gram models may reduce perplexity by $12.7\%$ compared to the current state-of-the-art using modified Kneser-Ney models on large data sets. 
Using small training data sets we observe even higher reductions of perplexity of up to $25.6\%$.
\end{enumerate}

The rest of the paper is organized as follows.
We start with reviewing related work in Section~\ref{sec:relwork}.
We will then introduce our generalized language models in Section~\ref{sec:notation}. 
After explaining the evaluation methodology and introducing the data sets in Section~\ref{sec:method} we will present the results of our evaluation in Section~\ref{sec:eval}.
In Section ~\ref{sec:discussion} we discuss why a generalized language model performs better than a standard language model.
Finally, in Section~\ref{sec:future} we summarize our findings and conclude with an overview of further interesting research challenges in the field of generalized language models.

\section{Related Work}\label{sec:relwork}

Work related to our generalized language model approach can be divided in two categories: various smoothing techniques for language models and approaches making use of skip $n$-grams.

Smoothing techniques for language models have a long history. 
Their aim is to overcome data sparsity and provide more reliable estimators---in particular for rare events.
The Good Turing estimator~\cite{good1953population}, deleted interpolation~\cite{P:PRP:1980:JelinekM}, Katz backoff~\cite{J:TASSP:1987:Katz} and Kneser-Ney smoothing~\cite{P:ICASSP:1995:KneserN} are just some of the approaches to be mentioned.
Common strategies of these approaches are to either backoff to lower order models when a higher order model lacks sufficient training data for good estimation, to interpolate between higher and lower order models or to interpolate with a prior distribution.
Furthermore, the estimation of the amount of unseen events from rare events aims to find the right weights for interpolation as well as for discounting probability mass from unreliable estimators and to retain it for unseen events.

The state of the art is a modified version of Kneser-Ney smoothing introduced in~\cite{J:CSL:1999:ChenG}. 
The modified version implements a recursive interpolation with lower order models, making use of different discount values for more or less frequently observed events.
This variation has been compared to other smoothing techniques on various corpora and has shown to outperform competing approaches.
We will review modified Kneser-Ney smoothing in Section~\ref{subsec:MKN} in more detail as we reuse some ideas to define our generalized language model.

Smoothing techniques which do not rely on using lower order models involve clustering~\cite{J:CL:1990:BrownSMP,J:CLS:1994:NeyEK}, i.e.\ grouping together similar words to form classes of words, as well as skip $n$-grams~\cite{J:CLS:1994:NeyEK,J:CSL:1993:HuangFHM}.
Yet other approaches make use of permutations of the word order in $n$-grams~\cite{schukat1995permugram,Tech:2001:Goodman}.

Skip $n$-grams are typically used to incorporate long distance relations between words. 
Introducing the possibility of gaps between the words in an $n$-gram allows for capturing word relations beyond the level of $n$ consecutive words without an exponential increase in the parameter space.
However, with their restriction on a subsequence of words, skip $n$-grams are also used as a technique to overcome data sparsity~\cite{Tech:2001:Goodman}.
In related work different terminology and different definitions have been used to describe skip $n$-grams.
Variations modify the number of words which can be skipped between elements in an $n$-gram as well as the manner in which the skipped words are determined (e.g.\ fixed patterns~\cite{Tech:2001:Goodman} or functional words~\cite{P:IJCNLP:2004:GaoS}).

The impact of various extensions and smoothing techniques for language models is investigated in~\cite{Tech:2001:Goodman,P:ICASSP:2000:Goodman}.
In particular, the authors compared Kneser-Ney smoothing, Katz backoff smoothing, caching, clustering, inclusion of higher order $n$-grams, sentence mixture and skip $n$-grams.
They also evaluated combinations of techniques, for instance, using skip $n$-gram models in combination with Kneser-Ney smoothing. 
The experiments in this case followed two paths: (1) interpolating a $5$-gram model with lower order distribution introducing a single gap and (2) interpolating higher order models with skip $n$-grams which retained only  combinations of two words.
Goodman reported on small data sets and in the best case a moderate improvement of cross entropy in the range of $0.02$ to $0.04$. 

In \cite{P:LREC:2006:GuthrieALG}, the authors investigated the increase of observed word combinations when including skips in $n$-grams.
The conclusion was that using skip $n$-grams is often more effective for increasing the number of observations than increasing the corpus size. 
This observation aligns well with our experiments.

\subsection{Review of Modified Kneser-Ney Smoothing}
\label{subsec:MKN}

We briefly recall modified Kneser-Ney Smoothing as presented in~\cite{J:CSL:1999:ChenG}.
Modified Kneser-Ney implements smoothing by interpolating between higher and lower order $n$-gram language models.
The highest order distribution is interpolated with lower order distribution as follows:

{\small
\begin{align}
\label{eq:smoothing:mod-kneser-ney-high}
P_{\text{MKN}}&(w_i|w_{i-n+1}^{i-1})=
&\frac{\text{max}\{ c(w_{i-n+1}^i)-D(c(w_{i-n+1}^i)),0\}}{c(w_{i-n+1}^{i-1})} 
&+\gamma_{high}(w_{i-n+1}^{i-1}){\hat P}_{\text{MKN}}(w_i|w_{i-n+2}^{i-1})
\end{align}
}%

where $c(w_{i-n+1}^i)$ provides the frequency count that sequence $w_{i-n+1}^i$ occurs in training data, $D$ is a discount value (which depends on the frequency of the sequence) and $\gamma_{high}$ depends on $D$ and is the interpolation factor to mix in the lower order distribution\footnote{The factors $\gamma$ and $D$ are quite technical and lengthy. As they do not play a significant role for understanding our novel approach we refer to Appendix~\ref{app:kmn} for details.}.
Essentially, interpolation with a lower order model corresponds to leaving out the first word in the considered sequence.
The lower order models are computed differently using the notion of continuation counts rather than absolute counts: 

{\small
\begin{align}
{\hat P}_{\text{MKN}}&(w_i| (w_{i-n+1}^{i-1}))=
&\frac{\text{max}\{ N_{1+}(\bullet w_{i-n+1}^i)-D(c(w_{i-n+1}^i)),0\}}{N_{1+}(\bullet w_{i-n+1}^{i-1}\bullet)}
&+\gamma_{mid}(w_{i-n+1}^{i-1}) {\hat P}_{\text{MKN}}(w_i|w_{i-n+2}^{i-1}))
\end{align}
}

where the continuation counts are defined as $N_{1+}(\bullet w_{i-n+1}^i) = |\{w_{i-n}: c(w_{i-n}^{i})>0\}|$, i.e. the number of different words which precede the sequence $w_{i-n+1}^i$.
The term $\gamma_{mid}$ is again an interpolation factor which depends on the  discounted probability mass $D$ in the first term of the formula.

\section{Generalized Language Models}\label{sec:notation}

\subsection{Notation for Skip $n$-gram with $k$ Skips}
We express skip $n$-grams using an operator notation.
The operator $\skp{i}$ applied to an $n$-gram removes the word at the $i$-th position.
For instance: $\skp{3} w_1 w_2 w_3 w_4 = w_1 w_2 \_ w_4 $, where $\_$ is used as wildcard placeholder to indicate a removed word.
The wildcard operator allows for larger number of matches. For instance, when $c(w_1 w_2 w_{3a} w_4) = x$ and $c(w_1 w_2 w_{3b} w_4)=y$ then $c(w_1 w_2 \_ w4)\geq x+y$ since at least the two sequences $w_1 w_2 w_{3a} w_4$ and  $w_1 w_2 w_{3b} w_4$ match the sequence $w_1 w_2 \_ w_4$.
In order to align with standard language models the skip operator applied to the first word of a sequence will remove the word instead of introducing a wildcard.
In particular the equation $\skp{1} w_{i-n+1}^{i} =  w_{i-n+2}^{i}$ holds where the right hand side is the subsequence of $w_{i-n+1}^{i}$ omitting the first word.
We can thus formulate the interpolation step of modified Kneser-Ney smoothing using our notation as ${\hat P}_{\text{MKN}}(w_i|w_{i-n+2}^{i-1}) = {\hat P}_{\text{MKN}}(w_i| \skp{1} w_{i-n+1}^{i-1})$.

Thus, our skip $n$-grams correspond to $n$-grams of which we only use $k$ words, after having applied 
the skip operators $\skp{i_1}\dots\skp{i_{n-k}}$

\subsection{Generalized Language Model}

Interpolation with lower order models is motivated by the problem of data sparsity in higher order models.
However, lower order models omit only the first word in the local context, which might not necessarily be the cause for the overall $n$-gram to be rare.
This is the motivation for our generalized language models to not only interpolate with one lower order model, where the first word in a sequence is omitted, but also with all other skip $n$-gram models, where one word is left out.
Combining this idea with modified Kneser-Ney smoothing leads to a formula similar to~(\ref{eq:smoothing:mod-kneser-ney-high}). 

{\small
\begin{align}
\label{eq:smoothing:glm-mkn-high}
P_{\text{GLM}}&(w_i|w_{i-n+1}^{i-1})= 
&\frac{\text{max}\{ c(w_{i-n+1}^i)-D(c(w_{i-n+1}^i)),0\}}{c(w_{i-n+1}^{i-1})}
&+\gamma_{high}(w_{i-n+1}^{i-1}) \sum_{j=1}^{n-1} \frac{1}{n\!-\!1} {\hat P}_{\text{GLM}}(w_i| \skp{j} w_{i-n+1}^{i-1})
\end{align}
}%

The difference between formula (\ref{eq:smoothing:mod-kneser-ney-high}) and formula (\ref{eq:smoothing:glm-mkn-high}) is the way in which lower order models are interpolated.

Note, the sum over all possible positions in the context $w_{i-n+1}^{i-1}$ for which we can skip a word and the according lower order models $P_{\text{GLM}}(w_i| \skp{j} (w_{i-n+1}^{i-1}))$.
We give all lower order models the same weight $\frac{1}{n-1}$.

The same principle is recursively applied in the lower order models in which some words of the full $n$-gram are already skipped.
As in modified Kneser-Ney smoothing we use continuation counts for the lower order models, incorporating the skip operator also for these counts.
Incorporating this directly into modified Kneser-Ney smoothing leads in the second highest model to:

{\small
\begin{align}
{\hat P}_{\text{GLM}}&(w_i| \skp{j}(w_{i\!-\!n\!+\!1}^{i-1}))=
&\frac{\text{max}\{ N_{1+}(\skp{j}(w_{i\!-\!n}^i))-D(c(\skp{j}(w_{i\!-\!n\!+\!1}^i))),0\}}{N_{1+}(\skp{j}(w_{i\!-\!n\!+\!1}^{i\!-\!1})\bullet)}
&+\!\gamma_{mid}(\skp{j}(w_{i\!-\!n\!+\!1}^{i\!-\!1}))\!\sum_{k=1 \atop k\neq j}^{n-1} \frac{1}{n\!-\!2}{\hat P}_{\text{GLM}}(w_i|\skp{j}\skp{k}(w_{i\!-\!n\!+\!1}^{i\!-\!1})) \nonumber
\end{align}
}%

Given that we skip words at different positions, we have to extend the notion of the count function and the continuation counts. 
The count function applied to a skip $n$-gram is given by  $c(\skp{j}(w_{\!i-\!n}^i))\!=\!\sum_{w_{j}}c(w_{\!i-\!n}^i)$, i.e.\ we aggregate the count information over all words which fill the gap in the $n$-gram.
Regarding the continuation counts we define:

{\small
\begin{align}
N_{1+}(\skp{j}(w_{\!i-\!n}^i)) & = & |\{w_{i\!-\!n\!+\!j\!-\!1}\!:\!c(w_{i\!-\!n}^{i})\!>\!0\}| \\
N_{1+}(\skp{j}(w_{i\!-\!n}^{i\!-\!1})\bullet) & = & |\{(w_{i\!-\!n\!+\!j\!-\!1},w_i)\!:\!c(w_{i\!-\!n}^{i})\!>\!0\}| 
\end{align}
}%

As lowest order model we use---just as done for traditional modified Kneser-Ney~\cite{J:CSL:1999:ChenG}---a unigram model interpolated with a uniform distribution for unseen words.

\begin{figure}[btph]
\includegraphics[width=0.9\columnwidth]{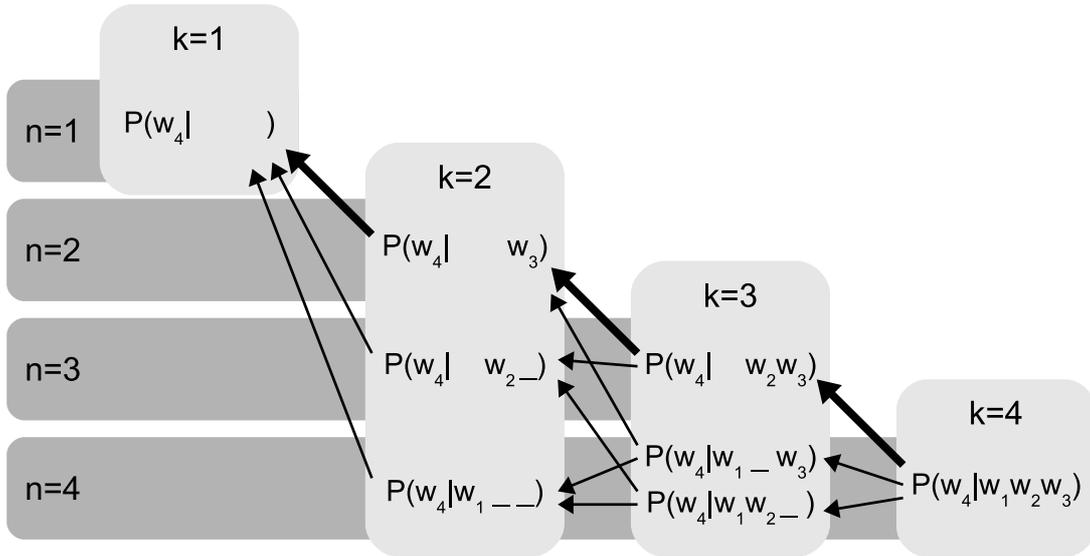} 
\centering
\caption{Interpolation of models of different order and using skip patterns. 
The value of $n$ indicates the length of the raw $n$-grams necessary for computing the model, the value of $k$ indicates the number of words actually used in the model. 
The wild card symbol \_ marks skipped words in an $n$-gram.
The arrows indicate how a higher order model is interpolated with lower order models which skips one word. 
The bold arrows correspond to interpolation of models in traditional modified Kneser-Ney smoothing. 
The lighter arrows illustrate the additional interpolations introduced by our generalized language models.
}
\label{fig:glmSmoothing}
\end{figure}

The overall process is depicted in Figure~\ref{fig:glmSmoothing}, illustrating how the higher level models are recursively smoothed with several lower order ones.

\section{Experimental Setup and Data Sets}\label{sec:method}

To evaluate the quality of our generalized language models we empirically compare their ability to explain sequences of words.
To this end we use text corpora, split them into test and training data, build language models as well as generalized language models over the training data and apply them on the test data.
We employ established metrics, such as cross entropy and perplexity.
In the following we explain the details of our experimental setup.

\subsection{Data Sets}
For evaluation purposes we employed eight different data sets.
The data sets cover different domains and languages.
As languages we considered English (\emph{en}), German (\emph{de}), French (\emph{fr}), and Italian (\emph{it}).
As general domain data set we used the full collection of articles from Wikipedia (\emph{wiki}) in the corresponding languages.
The download dates of the dumps are displayed in Table~\ref{tab:wikidates}.
\begin{table}[h]
\centering
  \begin{tabularx}{0.47\textwidth}{c|c|c|c}
   de & en & fr & it \\
    \hline
    Nov 22\textsuperscript{nd} & Nov 04\textsuperscript{th} & Nov 20\textsuperscript{th} & Nov 25\textsuperscript{th} \\
  \end{tabularx}
\caption{Download dates of Wikipedia snapshots in November 2013.}
\label{tab:wikidates}
\end{table}

Special purpose domain data are provided by the multi-lingual JRC-Acquis corpus of legislative texts (\emph{JRC})~\cite{P:LREC:2006:SteinbergerPWI}.
Table~\ref{tab:dataSetStats} gives an overview of the data sets and provides some simple statistics of the covered languages and the size of the collections.

\begin{table}[h]
\centering
\begin{tabularx}{0.45\textwidth}{Y | c c }
\toprule
\multicolumn{1}{c}{} & \multicolumn{2}{c}{Statistics} \\
Corpus & total words & unique words  \\ 
 &  in Mio. & in Mio. \\ 
  \midrule
wiki-de & 579 & 9.82  \\ 
JRC-de & 30.9 & 0.66  \\ 
  \hline
wiki-en & 1689 & 11.7  \\ 
JRC-en & 39.2 & 0.46 \\ 
  \hline
wiki-fr & 339 & 4.06 \\ 
JRC-fr & 35.8 & 0.46 \\ 
  \hline
wiki-it & 193 & 3.09 \\ 
JRC-it & 34.4 & 0.47 \\ 
\bottomrule
\end{tabularx} 
\caption{Word statistics and size of of evaluation corpora} 
\label{tab:dataSetStats} 
\end{table}

The data sets come in the form of structured text corpora which we cleaned from markup and tokenized to generate word sequences.
We filtered the word tokens by removing all character sequences which did not contain any letter, digit or common punctuation marks.
Eventually, the word token sequences were split into word sequences of length $n$ which provided the basis for the training and test sets for all algorithms.
Note that we did not perform case-folding nor did we apply stemming algorithms to normalize the word forms.
Also, we did our evaluation using case sensitive training and test data. 
Additionally, we kept all tokens for named entities such as names of persons or places.

\subsection{Evaluation Methodology}
All data sets have been randomly split into a training and a test set on a sentence level.
The training sets consist of 80\% of the sentences, which have been used to derive $n$-grams, skip $n$-grams and corresponding continuation counts for values of $n$ between 1 and 5. 
Note that we have trained a prediction model for each data set individually.
From the remaining 20\% of the sequences we have randomly sampled a separate set of $100,000$ sequences of $5$ words each. 
These test sequences have also been shortened to sequences of length $3$, and $4$ and provide a basis to conduct our final experiments to evaluate the performance of the different algorithms. 

We learnt the generalized language models on the same split of the training corpus as the standard language model using modified Kneser-Ney smoothing and we also used the same set of test sequences for a direct comparison.
To ensure rigour and openness of research the data set for training as well as the test sequences and the entire source code is open source. \footnote{http://west.uni-koblenz.de/Research} \footnote{https://github.com/renepickhardt/generalized-language-modeling-toolkit} \footnote{http://glm.rene-pickhardt.de}
We compared the probabilities of our language model implementation (which is a subset of the generalized language model) using KN as well as MKN smoothing with the Kyoto Language Model Toolkit \footnote{http://www.phontron.com/kylm/}. Since we got the same results for small $n$ and small data sets we believe that our implementation is correct.

In a second experiment we have investigated the impact of the size of the training data set. 
The wikipedia corpus consists of $1.7$ bn. words. 
Thus, the $80\%$ split for training consists of $1.3$ bn. words.
We have iteratively created smaller training sets by decreasing the split factor by an order of magnitude.
So we created $8\%$ / $92\%$ and $0.8\%$ / $99.2\%$ split, and so on.
We have stopped at the $0.008\% / 99.992\%$ split as the training data set in this case consisted of less words than our 100k test sequences which we still randomly sampled from the test data of each split.
Then we trained a generalized language model as well as a standard language model with modified Kneser-Ney smoothing on each of these samples of the training data.
Again we have evaluated these language models on the same random sample of $100,000$ sequences as mentioned above.

\subsection{Evaluation Metrics}
As evaluation metric we use \emph{perplexity}: a standard measure in the field of language models ~\cite{Manning:1999:LM}. 
First we calculate the \emph{cross entropy} of a trained language model given a test set using 

\begin{equation}
H(P_{\tt{alg}}) = - \sum_{s\in T}P_{\tt{MLE}}(s) \cdot \log_2{P_{\tt{alg}}(s)}
\end{equation}

Where $P_{\tt{alg}}$ will be replaced by the probability estimates provided by our generalized language models and the estimates of a language model using modified Kneser-Ney smoothing. 
 $P_{\tt{MLE}}$, instead, is a maximum likelihood estimator of the test sequence to occur in the test corpus.
Finally, $T$ is the set of test sequences.
The perplexity is defined as: 

\begin{equation}
\textit{Perplexity}(P_{\tt{alg}}) = 2^{H(P_{\tt{alg}})}
\end{equation}

Lower perplexity values indicate better results.

\section{Results}\label{sec:eval}
\subsection{Baseline}
As a baseline for our generalized language model (GLM) we have trained standard language models using modified Kneser-Ney Smoothing (MKN). 
These models have been trained for model lengths $3$ to $5$.
For unigram and bigram models MKN and GLM are identical.

\subsection{Evaluation Experiments}
The perplexity values for all data sets and various model orders can be seen in Table~\ref{tab:fullPerplexity}. 
In this table we also present the relative reduction of perplexity in comparison to the baseline. 

\begin{table}[tbhp]
\centering
\begin{tabularx}{0.45\textwidth}{Y | c c c}
\toprule
\multicolumn{1}{c}{} & \multicolumn{3}{c}{model length} \\
Experiments & $n=3$ & $n=4$ & $n=5$  \\ 
  \midrule						
wiki-de MKN & 1074.1 & 778.5 & 597.1   \\ 
wiki-de GLM & \textbf{1031.1} & \textbf{709.4} & \textbf{521.5}   \\ 
rel. change & 4.0\% & 8.9\% & 12.7\%   \\ 
  \midrule						
JRC-de MKN & 235.4 & 138.4 & 94.7   \\ 
JRC-de GLM & \textbf{229.4} & \textbf{131.8} & \textbf{86.0}   \\ 
rel. change & 2.5\% & 4.8\% & 9.2\%   \\ 
  \hline						
wiki-en MKN & 586.9 & 404 & 307.3   \\ 
wiki-en GLM & \textbf{571.6} & \textbf{378.1} & \textbf{275}  \\ 
rel. change & 2.6\% & 6.1\% & 10.5\%   \\ 
  \midrule						
JRC-en MKN & 147.2 & 82.9 & 54.6   \\ 
JRC-en GLM & \textbf{145.3} & \textbf{80.6} & \textbf{52.5}  \\ 
rel. change & 1.3\% & 2.8\% & 3.9\%   \\ 
  \hline						
wiki-fr MKN & 538.6 & 385.9 & 298.9   \\ 
wiki-fr GLM & \textbf{526.7} & \textbf{363.8} & \textbf{272.9}  \\ 
rel. change & 2.2\% & 5.7\% & 8.7\%   \\ 
  \midrule						
JRC-fr MKN & 155.2 & 92.5 & 63.9   \\ 
JRC-fr GLM & \textbf{153.5} & \textbf{90.1} & \textbf{61.7}  \\ 
rel. change & 1.1\% & 2.5\% & 3.5\%   \\ 
  \hline						
wiki-it MKN & 738.4 & 532.9 & 416.7   \\ 
wiki-it GLM & \textbf{718.2} & \textbf{500.7} & \textbf{382.2} \\ 
rel. change & 2.7\% & 6.0\% & 8.3\%   \\ 
  \midrule						
JRC-it MKN & 177.5 & 104.4 & 71.8 \\ 
JRC-it GLM & \textbf{175.1} & \textbf{101.8} & \textbf{69.6}  \\ 
rel. change & 1.3\% & 2.6\% & 3.1\%  \\ 
\bottomrule
\end{tabularx} 
\caption{Absolute perplexity values and relative reduction of perplexity from MKN to GLM on all data sets for models of order $3$ to $5$} 
\label{tab:fullPerplexity} 
\end{table}

As we can see, the GLM clearly outperforms the baseline for all model lengths and data sets.
In general we see a larger improvement in performance for models of higher orders ($n=5$). 
The gain for 3-gram models, instead,  is negligible.
For German texts the increase in performance is the highest ($12.7\%$) for a model of order $5$.
We also note that GLMs seem to work better on broad domain text rather than special purpose text as the reduction on the wiki corpora is constantly higher than the reduction of perplexity on the JRC corpora.

\begin{figure*}[tbhp]
  \centering
    \includegraphics[width=0.9\textwidth]{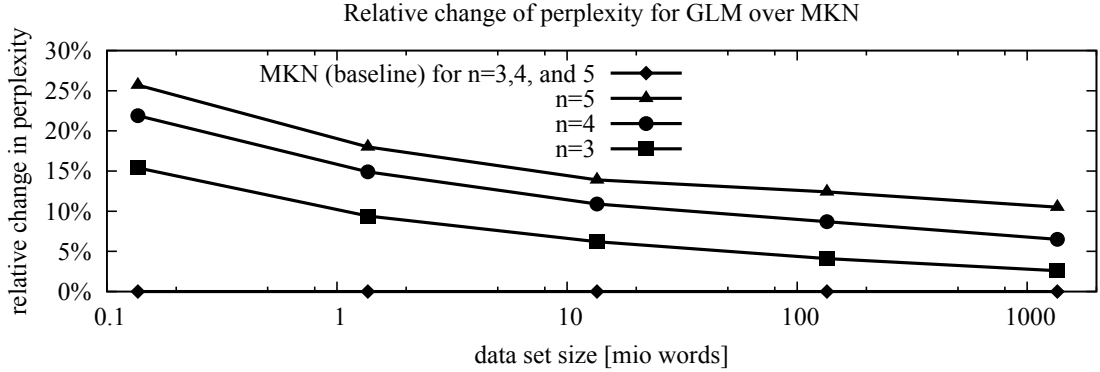}
    \caption{Variation of the size of the training data on 100k test sequences on the English Wikipedia data set with different model lengths for GLM.}
  \label{fig:GLMCorpusSize}
\end{figure*}

We made consistent observations in our second experiment where we iteratively shrank the size of the training data set.
We calculated the relative reduction in perplexity from MKN to GLM for various model lengths and the different sizes of the training data.
The results for the English Wikipedia data set are illustrated in Figure~\ref{fig:GLMCorpusSize}.

We see that the GLM performs particularly well on small training data. 
As the size of the training data set becomes smaller (even smaller than the evaluation data), the GLM achieves a reduction of perplexity of up to $25.7\%$ compared to language models with modified Kneser-Ney smoothing on the same data set.
The absolute perplexity values for this experiment are presented in Table~\ref{tab:fullPerplexityDataSize}.

\begin{table}[tbhp]
\centering
\begin{tabularx}{0.45\textwidth}{Y | c c c}
\toprule
\multicolumn{1}{c}{} & \multicolumn{3}{c}{model length} \\
Experiments & $n=3$ & $n=4$ & $n=5$  \\ 
  \midrule						
$80\%$ MKN & 586.9 & 404 & 307.3   \\ 
$80\%$ GLM & \textbf{571.6} & \textbf{378.1} & \textbf{275}  \\ 
rel. change & 2.6\% & 6.5\% & 10.5\%  \\
  \midrule						
$8\%$ MKN & 712.6 & 539.8 & 436.5   \\ 
$8\%$ GLM & \textbf{683.7} & \textbf{492.8} & \textbf{382.5}  \\ 
rel. change & 4.1\% & 8.7\% & 12.4\%  \\
  \midrule						
$0.8\%$ MKN & 894.0 & 730.0 & 614.1   \\ 
$0.8\%$ GLM & \textbf{838.7} & \textbf{650.1} & \textbf{528.7}  \\ 
rel. change & 6.2\% & 10.9\% & 13.9\%  \\
  \midrule						
$0.08\%$ MKN & 1099.5 & 963.8 & 845.2   \\ 
$0.08\%$ GLM & \textbf{996.6} & \textbf{820.7} & \textbf{693.4}  \\ 
rel. change & 9.4\% & 14.9\% & 18.0\%  \\
  \midrule						
$0.008\%$ MKN & 1212.1 & 1120.5 &  1009.6  \\ 
$0.008\%$ GLM & \textbf{1025.6} & \textbf{875.5} & \textbf{750.3}  \\ 
rel. change & 15.4\% & 21.9\% & 25.7\%  \\
\bottomrule
\end{tabularx} 
\caption{Absolute perplexity values and relative reduction of perplexity from MKN to GLM on shrunk training data sets for the English Wikipedia for models of order $3$ to $5$} 
\label{tab:fullPerplexityDataSize} 
\end{table}

Our theory as well as the results so far suggest that the GLM performs particularly well on sparse training data. 
This conjecture has been investigated in a last experiment. 
For each model length we have split the test data of the largest English Wikipedia corpus into two disjoint evaluation data sets. 
The data set \textit{unseen} consists of all test sequences which have never been observed in the training data.
The set \textit{observed} consists only of test sequences which have been observed at least once in the training data. 
Again we have calculated the perplexity of each set. 
For reference, also the values of the \text{complete} test data set are shown in Table~\ref{tab:sparsity}.

\begin{table}[tbhp]
\centering
\begin{tabularx}{0.47\textwidth}{Y | c c c}
\toprule
\multicolumn{1}{c}{} & \multicolumn{3}{c}{model length} \\
Experiments & $n=3$ & $n=4$ & $n=5$  \\ 
  \midrule						
MKN\textsuperscript{complete} & 586.9 & 404 & 307.3   \\ 
GLM\textsuperscript{complete} & \textbf{571.6} & \textbf{378.1} & \textbf{275}  \\ 
rel. change & 2.6\% & 6.5\% & 10.5\%   \\ 
  \midrule						
MKN\textsuperscript{unseen} & 14696.8 & 2199.8 & 846.1   \\ 
GLM\textsuperscript{unseen} & \textbf{13058.7} & \textbf{1902.4} & \textbf{714.4}  \\ 
rel. change & 11.2\% & 13.5\% & 15.6\%   \\ 
  \midrule						
MKN\textsuperscript{observed} & \textbf{220.2} & \textbf{88.0} & \textbf{43.4}  \\ 
GLM\textsuperscript{observed} & 220.6 & 88.3 & 43.5   \\ 
rel. change & $-0.16\%$ & $-0.28\%$ & $-0.15\%$   \\ 
\bottomrule
\end{tabularx} 
\caption{Absolute perplexity values and relative reduction of perplexity from MKN to GLM for the complete and split test file into observed and unseen sequences for models of order $3$ to $5$. The data set is the largest English Wikipedia corpus.} 
\label{tab:sparsity} 
\end{table}

As expected we see the overall perplexity values rise for the \textit{unseen} test case and decline for the \textit{observed} test case. 
More interestingly we see that the relative reduction of perplexity of the GLM over MKN increases from $10.5\%$ to $15.6\%$ on the \textit{unseen} test case. 
This indicates that the superior performance of the GLM on small training corpora and for higher order models indeed comes from its good performance properties with regard to sparse training data.
It also confirms that our motivation to produce lower order $n$-grams by omitting not only the first word of the local context but systematically all words has been fruitful.
However, we also see that for the \textit{observed} sequences the GLM performs slightly worse than MKN.
For the observed cases we find the relative change to be negligible.

\section{Discussion}\label{sec:discussion}
In our experiments we have observed an improvement of our generalized language models over classical language models using Kneser-Ney smoothing.
The improvements have been observed for different languages, different domains as well as different sizes of the training data.
In the experiments we have also seen that the GLM performs well in particular for small training data sets and sparse data, encouraging our initial motivation.
This feature of the GLM is of particular value, as data sparsity becomes a more and more immanent problem for higher values of $n$.
This known fact is underlined also by the statistics shown in Table~\ref{tab:percentage-enwiki-complete}.
The fraction of total $n$-grams which appear only once in our Wikipedia corpus increases for higher values of $n$.
However, for the same value of $n$ the skip $n$-grams are less rare.
Our generalized language models leverage this additional information to obtain more reliable estimates for the probability of word sequences.

\begin{table}[ht]
\centering
\begin{tabularx}{0.4\textwidth}{Y | c | c}
\toprule

    \textbf{$w_1^n$}&\textbf{total}&\textbf{unique}\\
    \midrule
      $w_1$&$0.5\%$&$64.0\%$\\
    \midrule
      $w_1w_2$&$5.1\%$&$68.2\%$\\
      $w_1\_w_3$&$8.0\%$&$79.9\%$\\
      $w_1\_\_w_4$&$9.6\%$&$72.1\%$\\
      $w_1\_\_\_w_5$&$10.1\%$&$72.7\%$\\
    \midrule
      $w_1w_2w_3$&$21.1\%$&$77.5\%$\\
      $w_1\_w_3w_4$&$28.2\%$&$80.4\%$\\
      $w_1w_2\_w_4$&$28.2\%$&$80.7\%$\\
      $w_1\_\_w_4w_5$&$31.7\%$&$81.9\%$\\
      $w_1\_w_3\_w_5$&$35.3\%$&$83.0\%$\\
      $w_1w_2\_\_w_5$&$31.5\%$&$82.2\%$\\
    \midrule
      $w_1w_2w_3w_4$&$44.7\%$&$85.4\%$\\
      $w_1\_w_3w_4w_5$&$52.7\%$&$87.6\%$\\
      $w_1w_2\_w_4w_5$&$52.6\%$&$88.0\%$\\
      $w_1w_2w_3\_w_5$&$52.3\%$&$87.7\%$\\
    \midrule
      $w_1w_2w_3w_4w_5$&$64.4\%$&$90.7\%$\\
    \bottomrule
  \end{tabularx}
  \caption{Percentage of generalized $n$-grams which occur only once in the English Wikipedia corpus. Total means a percentage relative to the total amount of sequences. Unique means a percentage relative to the amount of unique sequences of this pattern in the data set.}
  \label{tab:percentage-enwiki-complete}
\end{table}

Beyond the general improvements there is an additional path for benefitting from generalized language models.
As it is possible to better leverage the information in smaller and sparse data sets, we can build smaller models of competitive performance.
For instance, when looking at Table~\ref{tab:fullPerplexityDataSize} we observe the $3$-gram MKN approach on the full training data set to achieve a perplexity of $586.9$.
This model has been trained on $7$ GB of text and the resulting model has a size of $15$ GB and $742$ Mio.\ entries for the count and continuation count values. 
Looking for a GLM with comparable but better performance we see that the $5$-gram model trained on $1\%$ of the training data has a perplexity of $528.7$.
This GLM model has a size of $9.5$ GB and contains only $427$ Mio. entries.
So, using a far smaller set of training data we can build a smaller model which still demonstrates  a competitive performance.   

\section{Conclusion and Future Work}\label{sec:future}
\subsection{Conclusion}
We have introduced a novel generalized language model as the systematic combination of skip $n$-grams and modified Kneser-Ney smoothing. 
The main strength of our approach is the combination of a simple and elegant idea with an an empirically convincing result.
Mathematically one can see that the GLM includes the standard language model with modified Kneser-Ney smoothing as a sub model and is consequently a real generalization. 

In an empirical evaluation, we have demonstrated that for higher orders the GLM outperforms MKN for all test cases.
The relative improvement in perplexity is up to $12.7\%$ for large data sets.
GLMs also performs particularly well on small and sparse sets of training data.
On a very small training data set we observed a reduction of perplexity by $25.7\%$.
Our experiments underline that the generalized language models overcome in particular the weaknesses of  modified Kneser-Ney smoothing on sparse training data.

\subsection{Future work}
A desirable extension of our current definition of GLMs will be the combination of different lower lower order models in our generalized language model using different weights for each model.
Such weights can be used to model the statistical reliability of the different lower order models.
The value of the weights would have to be chosen according to the probability or counts of the respective skip $n$-grams.

Another important step that has not been considered yet is compressing and indexing of generalized language models to improve the performance of the computation and be able to store them in main memory.
Regarding the scalability of the approach to very large data sets we intend to apply the Map Reduce techniques from~\cite{P:ACL:2013:HeafieldPCK} to our generalized language models in order to have a more scalable calculation. 

This will open the path also to another interesting experiment.
Goodman~\cite{Tech:2001:Goodman} observed that increasing the length of $n$-grams in combination with modified Kneser-Ney smoothing did not lead to improvements for values of $n$ beyond 7.
We believe that our generalized language models could still benefit from such an increase. They suffer less from the sparsity of long $n$-grams and can overcome this sparsity when interpolating with the lower order skip $n$-grams while benefiting from the larger context.

Finally, it would be interesting to see how applications of language models---like next word prediction, machine translation, speech recognition, text classification, spelling correction, e.g.---benefit from the better performance of generalized language models.

\section*{Acknowledgements}
We would like to thank Heinrich Hartmann for a fruitful discussion regarding notation of the skip operator for $n$-grams.
The research leading to these results has received funding from the European Community's Seventh Framework Programme (FP7/2007-2013), REVEAL (Grant agree number 610928).

\bibliographystyle{plain}
\bibliography{main}

\appendix 

\section{Discount Values and Weights in Modified Kneser Ney}
\label{app:kmn}

The discount value $D(c)$ used in formula~(\ref{eq:smoothing:mod-kneser-ney-high}) is defined as~\cite{J:CSL:1999:ChenG}:
{\small
\begin{equation}\label{eq:smoothing:mod-kneser-ney-d}
  D(c)=
  \begin{cases}\ 0 & \text{if}\ c=0 \\
    D_1 & \text{if}\ c=1 \\
    D_2 & \text{if}\ c=2 \\
    D_{3+} & \text{if}\ c>2 \\
  \end{cases}
\end{equation}
}%
The discounting values $D_1$, $D_2$, and $D_{3+}$ are defined as \cite{chen1998empirical}
{\small
\begin{subequations}\label{eq:smoothing:mod-kneser-ney-ds}
  \begin{align}
    D_1=1-2Y\frac{n_2}{n_1}\\
    D_2=2-3Y\frac{n_3}{n_2}\\
    D_{3+}=3-4Y\frac{n_4}{n_3}
  \end{align}
\end{subequations}
}%
with $Y=\frac{n_1}{n_1+n_2}$ and $n_i$ is the total number of $n$-grams which appear exactly $i$ times in the training data.
The weight $\gamma_{high}(w_{i-n+1}^{i-1})$ is defined as:
{\small
\begin{align}\label{eq:smoothing:mod-kneser-ney-gamma-high2}
\gamma_{high}&(w_{i\!-\!n\!+\!1}^{i\!-\!1})=
\frac{D_1N_1(w_{i\!-\!n\!+\!1}^{i\!-\!1}\!\bullet)\!+\!D_2N_2(w_{i\!-\!n\!+\!1}^{i\!-\!1}\!\bullet)\!+\!D_{3+}N_{3+}(w_{i\!-\!n\!+\!1}^{i\!-\!1}\!\bullet)}{c(w_{i\!-\!n\!+\!1}^{i\!-\!1})}
\end{align}
}%
And the weight $\gamma_{mid}(w_{i-n+1}^{i-1})$ is defined as:
{\small
\begin{align}\label{eq:smoothing:mod-kneser-ney-gamma-low1}
\gamma_{mid}&(w_{i\!-\!n\!+\!1}^{i\!-\!1})=&\frac{D_1N_1(w_{i\!-\!n\!+\!1}^{i\!-\!1}\!\bullet)\!+\!D_2N_2(w_{i\!-\!n\!+\!1}^{i\!-\!1}\!\bullet)\!+\!D_{3+}N_{3+}(w_{i\!-\!n\!+\!1}^{i\!-\!1}\!\bullet)}{N_{1+}(\bullet\! w_{i\!-\!n\!+\!1}^{i\!-\!1}\bullet)}
\end{align}
}%
where $N_1(w_{i-n+1}^{i-1}\bullet)$, $N_2(w_{i-n+1}^{i-1}\bullet)$, and $N_{3+}(w_{i-n+1}^{i-1}\bullet)$ are analogously defined to $N_{1+}(w_{i-n+1}^{i-1}\bullet)$.

\end{document}